\documentclass{article}
\usepackage{spconf,amsmath,graphicx}
\usepackage{amssymb}
\usepackage{microtype}
\usepackage{setspace}
\usepackage{siunitx}
\usepackage{booktabs}
\usepackage{multirow}
\usepackage{nicefrac}
\usepackage{tabularx}
\newcolumntype{Y}{>{\centering\arraybackslash}X}
\newcolumntype{L}{>{\raggedright\arraybackslash}X}
\newcolumntype{n}{>{\hsize=0.35\hsize}L}
\newcolumntype{t}{>{\hsize=1.56\hsize}L}
\newcolumntype{u}{>{\hsize=1.03\hsize}Y}
\usepackage[hidelinks]{hyperref}

\newcommand{\bftab}{\fontseries{b}\selectfont}
\newcommand{\STAB}[1]{\begin{tabular}{@{}c@{}}#1\end{tabular}}
\newcommand{\Rn}[1]{\expandafter{\romannumeral #1\relax}}
\usepackage[pages=some]{background}
\SetBgContents{\parbox{17.4cm}{\small
\textcopyright\ 2020 IEEE.  Personal use of this material is permitted.  Permission from IEEE must be obtained for all other uses, in any current or future media, including reprinting/republishing this material for advertising or promotional purposes, creating new collective works, for resale or redistribution to servers or lists, or reuse of any copyrighted component of this work in other works.}}
\SetBgScale{1}
\SetBgAngle{0}
\SetBgOpacity{1}
\SetBgColor{black}
\SetBgPosition{current page.north west}
\SetBgHshift{10.8cm}
\SetBgVshift{-1.8cm}
\title{
Discovering Salient Anatomical Landmarks by Predicting Human Gaze
}
\name{
R.\ Droste\textsuperscript{1},
P.\ Chatelain\textsuperscript{1},
L.\ Drukker\textsuperscript{2},
H.\ Sharma\textsuperscript{1},
A.\ T.\ Papageorghiou\textsuperscript{2},
J.\ A.\ Noble\textsuperscript{1}
}
\address{
\textsuperscript{1}Department of Engineering Science, \mbox{University of Oxford}, Oxford, UK\\
\textsuperscript{2}Nuffield Department of Women's \& Reproductive Health, \mbox{University of Oxford, Oxford, UK}
}
\begin{document}
\maketitle              
\begin{abstract}
Anatomical landmarks are a crucial prerequisite for many medical imaging tasks.
Usually, the set of landmarks for a given task is predefined by experts.
The landmark locations for a given image are then annotated manually or via machine learning methods trained on manual annotations.
In this paper, in contrast, we present a method to automatically discover and localize anatomical landmarks in medical images.
Specifically, we consider landmarks that attract the visual attention of humans, which we term \emph{visually salient landmarks}.
We illustrate the method for fetal neurosonographic images.
First, full-length clinical fetal ultrasound scans are recorded with live sonographer gaze-tracking.
Next, a convolutional neural network (CNN) is trained to predict the gaze point distribution (saliency map) of the sonographers on scan video frames.
The CNN is then used to predict saliency maps of unseen fetal neurosonographic images, and the landmarks are extracted as the local maxima of these saliency maps.
Finally, the landmarks are matched across images by clustering the landmark CNN features.
We show that the discovered landmarks can be used within affine image registration, with average landmark alignment errors between 4.1\% and 10.9\% of the fetal head long axis length.
\end{abstract}

\BgThispage

\section{Introduction}
An \emph{anatomical landmark} is ``a point of correspondence on each object that matches between and within populations'' and is assigned ``in some scientifically meaningful way'' \cite[p.~3]{Dryden2016}.
For brevity, we will refer to \emph{anatomical landmarks} simply as \emph{landmarks}.
The selection and localization of landmarks are essential steps for medical image analysis tasks such as image registration and shape analysis.
Usually, the set of landmarks for a given task is selected by experts a priori.
The landmark locations for a given image are then either annotated manually or via machine learning models trained on manual annotations.
However, when clinicians interpret images in practice based on experience, they may consider only a subset of the predefined landmarks, or use additional, unspecified landmarks.
Moreover, it might be desirable to automatically localize landmarks without the need for manual annotations.

\emph{Contribution.\quad}
In this work we overcome these limitations by presenting a method to automatically discover and localize anatomical landmarks.
Specifically, the method reveals landmarks that attract the visual attention of clinicians, which we term \emph{visually salient landmarks}.
The backbone of the proposed system is a CNN that is trained to predict the gaze-point distributions (saliency maps) of clinicians observing images from the domain of interest.
For modalities like ultrasound imaging, gaze-tracking data can be acquired during image acquisition with no additional expert time expenditure.
The trained CNN is then used to reveal visually salient landmarks on unseen images and to assign them semantic labels that can be used to match them across images.
To the best of our knowledge, this is the first work to present a method to automatically discover landmarks based on \emph{visual} saliency.

\emph{Related Work.\quad}
In previous work, \emph{saliency} is often used to refer to low-level features such as local entropy \cite{Kadir2001,GuorongWu2006}.
Moreover, mutually-salient landmarks based on Gabor attributes have been proposed for image registration \cite{Ou2011}.
Here, in contrast, we use \emph{visual saliency}, i.e., the predicted allocation of human visual attention based on gaze-tracking data, to discover \emph{anatomical landmarks}.
We apply the method to neurosonographic standard views in fetal anomaly ultrasound scans.
The landmarks for these standard views are defined by a set of international practice guidelines \cite{Salomon2011}.
A landmark detector has previously been developed but is trained on manual annotations of a pre-defined set of landmarks \cite{Yaqub2017}.
Moreover, gaze data has been used to support the detection of standard views in fetal ultrasound scans \cite{Cai2018c,Droste2019b}, but these works do not consider the problem of identifying landmarks.

\begin{figure*}[tb]
{
\scriptsize
\sffamily
\noindent\hspace{0.1\linewidth}
\begin{minipage}{0.8\textwidth}
\def\svgwidth{1.\textwidth}
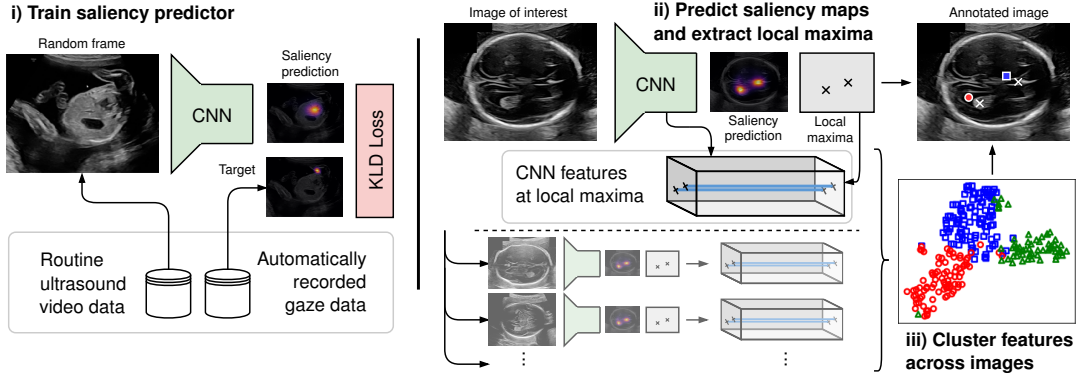
\end{minipage}
}
\caption{
Overview of the proposed method for the discovery and localization of visually salient landmarks.
}
\label{fig:method}
\end{figure*}

\section{Methods}
\label{sec:method}
\subsection{Data}
\label{sec:data}
The data were acquired as part of the PULSE (Perception Ultrasound by Learning Sonographic Experience) project, a prospective study of routine fetal ultrasound scans performed in all trimesters by sonographers and fetal medicine doctors at the maternity ultrasound unit, Oxford University Hospitals NHS Foundation Trust, Oxfordshire, United Kingdom.
The exams were performed on a GE Voluson E8 scanner (General Electric, USA) while the video signal of the machine monitor was recorded lossless at 30 Hz.
Operator gaze was simultaneously recorded at 90 Hz with a Tobii Eye Tracker 4C (Tobii, Sweden).
This study was approved by the UK Research Ethics Committee (Reference 18/WS/0051), and written informed consent was given by all participating pregnant women and operators.
In this paper, we use ultrasound video and corresponding gaze data of 212 second trimester scans acquired between May 2018 and February 2019.

We selected 90 scans to train the saliency predictor and used the remaining 122 scans to evaluate the landmark discovery method.
We considered the fetal neurosonographic standard views, i.e.,  the transventricular (TV) and the transcerebellar (TC) plane (first row in \autoref{fig:examples}).
On the TV plane the operators measure the head circumference (HC) and the lateral ventricle (LV). On the TC plane they measure the transcerebellar diameter (TCD), the nuchal fold and the cisterna magna.
The views are defined by the visibility of these structures as well as the appearance of the cavum septi pellucidi (CSP).
From the 122 ultrasound scans, we automatically extracted 143 TV and 124 TC plane images by performing optical character recognition on the machine's graphical interface.

\subsection{Visually Salient Landmark Discovery}
Visually salient anatomical landmarks are discovered in three steps (see \autoref{fig:method}):
i) training a CNN to predict the sonographer gaze point distributions (saliency maps) on random video frames of the routine fetal ultrasound scan data described above;
ii) predicting the visual saliency maps of the neurosonographic images and extracting the landmark locations as the local maxima of the saliency maps;
and iii) clustering the CNN feature vectors which correspond to the landmark locations.

\Rn{1})
To train the saliency predictor, we use the CNN architecture and training procedure detailed in previous work \cite{Droste2019b} (model \emph{Saliency-VAM}).
The precise architecture and training procedure are not repeated here as they are not essential for the proposed landmark discovery method.
The CNN takes ultrasound images of dimension 288$\times$244 as input and performs three two-fold down-sampling operations, which results in output saliency maps of dimensions $W_s$$\times$$H_s = 36$$\times$$28$.

\Rn{2})
Let $s_i: [1, W_s]\times [1, H_s]\cap \mathbb{Z}^2 \rightarrow [0, 1]$ be the function which, for an image with index $i=1,\dots,N_i$, maps each saliency map location to its predicted saliency value (i.e., the probability that the location
is gazed at).
The local maxima of this predicted saliency map are found with the scikit-image (https://scikit-image.org/) \emph{peak\_local\_max} algorithm.
The algorithm first applies a maximum filter
\begin{equation}
s_i^\mathrm{max}(x, y) := \max_{(x', y') \in [-d, d]^2\cap\mathbb{Z}^2} s_i(x + x', y + y')\;,
\end{equation}
where $d$ is the minimum distance of any two local maxima (empirically $d=2$).
The local maxima are then extracted as the points where the $s$ equals $s^\mathrm{max}$ and $s$ is above a threshold $t$ to suppress spurious maxima (empirically $t=0.1$):
\begin{equation}
\mathcal{M}_i := \left\lbrace (x, y) | s_i(x, y) = s_i^\mathrm{max}(x, y) \land s_i(x, y) \geq t\right\rbrace
\end{equation}
The landmark locations are obtained by fitting a 2D Gaussian peak to a 3$\times$3 neighborhood around the saliency map maxima.

\Rn{3})
Once the landmark locations are extracted, their correspondence across images is still unknown.
Recent work has shown that saliency predictors implicitly learn \emph{global} semantic features which are useful for image classification \cite{Droste2019b}.
Here, we hypothesize that saliency predictors can also be used to extract \emph{local} semantic features which allow automatic landmark classification.
Let $f_i: [1, W_s]\times [1, H_s]\cap \mathbb{Z}^2 \rightarrow \mathbb{R}^{N_f}$ be the function which, for image $i$, maps each location of the saliency map to the corresponding feature activations of the last CNN layer, where $N_f$ is the number of channels.
Then the set of all landmark feature vectors $\mathcal{F}$ across $N_i$ images is obtained as
\begin{equation}
\mathcal{F} := \bigcup_{i=1}^{N_i} \left\lbrace f_i(x, y) | (x, y) \in \mathcal{M}_i\right\rbrace\,.
\end{equation}
Finally, the feature vectors are classified via k-means clustering of $\mathcal{F}$.
The number of clusters is automatically selected by maximizing the \mbox{\emph{Silhouette Coefficient}} $\frac{1}{N_i}\Sigma_{i=1}^{N_i}\frac{b(i)-a(i)}{\max\lbrace a(i),b(i)\rbrace}$, where $a(i)$ is the mean intra-cluster distance and $b(i)$ the mean nearest-cluster distance of sample $i$ \cite{Rousseeuw1987}.

\subsection{Application to Image Registration}
In order to examine a simple practical use of the visually salient landmarks, we consider the task of aligning the standard view images.
For each plane, we use two landmarks to construct an affine transformation of optional horizontal flipping, translation, rotation and isotropic scaling.

\begin{figure}[!htb]
{
\centering
\tiny
\sffamily
\def\svgwidth{1.\columnwidth}
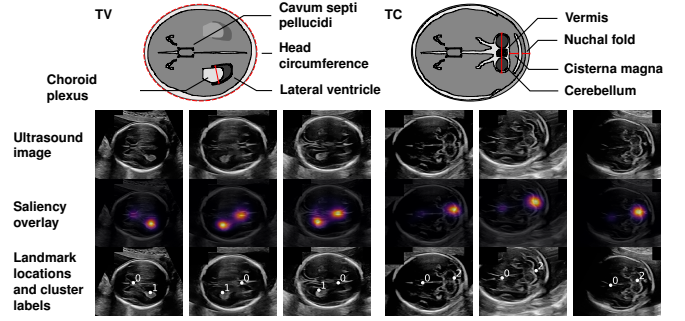
}
\caption{
Exemplary results of the visually salient landmark discovery method.
The top row illustrates the anatomy of the respective standard view, with biometric measurements highlighted in red \cite{NHS2018}.
The first row of the image grid shows exemplary neurosonographic images.
The second row shows an overlay of the predicted saliency map.
The third row shows the discovered landmarks with cluster labels.
}
\label{fig:examples}
\end{figure}

Consider the TV plane (the generalization to the TC plane is straightforward).
For image index $i$, let $C^{i} = \left(c_x^{i}, c_y^i\right) \in \mathbb{R}^2$ be the coordinates of the salient landmark corresponding to the CSP, and let $D^i = \left(d_x^i, d_y^i\right) \in \mathbb{R}^2$ be the coordinates of the landmark corresponding to the LV (or the cerebellum for the TC plane).
Let $j$ and $k$ be the indices of the source and target images to be aligned.
For a point $p=(p_x, p_y)$ on the images with width $W_i$, optional flipping of the $x$-coordinate is performed with the function
$f:\mathbb{R}\rightarrow\mathbb{R}$ with
\begin{equation}
f(p_x) = 
\begin{cases}
W_i - p_x &\text{if } \mathrm{sgn}(c_x^t - d_x^t) \neq \mathrm{sgn}(c_x^j - d_x^j)\\
p_x & \text{otherwise,}
\end{cases}
\end{equation}
which makes use of the fact that the horizontal ordering of the landmarks determines the orientation of the fetal head (see \autoref{fig:examples}).
Let $C^{j,f} = \left(f(c_x^j), c_y^j\right)$ and $D^{j,f} = \left(f(d_x^j), d_y^j\right)$ be the source image landmarks after optional horizontal flipping.
Next, the images are aligned with the translation vector $\mathbf{t} = (t_x, t_y) = \overrightarrow{C^{j,f} C^k}$, the isotropic scaling factor $\rho = \nicefrac{\|\overrightarrow{C^k D^k}\|}{\|\overrightarrow{C^{j,f} D^{j,f}}\|}$ and the rotation angle $\theta = \angle (\overrightarrow{C^{j,f} D^{j,f}}, \overrightarrow{C^k D^k})$, where the latter two operations are performed with center $C^k$.
The resulting affine transformation $\mathcal{T}^{j,k}:\mathbb{R}^2\rightarrow \mathbb{R}^2$ of a point $P^j = (p^j_x, p^j_y)$ on the source image to the estimated point $\hat{P}^k = (\hat{p}^k_x, \hat{p}^k_y)$ on the target image is
\begin{align}
\begin{bmatrix}
\hat{p}^k_x\\
\hat{p}^k_y\\
1\\
\end{bmatrix}
=&
\setlength{\arraycolsep}{4pt}
\begin{bmatrix}
\alpha & \beta & (1-\alpha)\,c_x^k - \beta\,c_y^k\\
-\beta & \alpha & \beta\,c_x^k + (1-\alpha)\,c_y^k\\
0&0&1\\
\end{bmatrix}
\begin{bmatrix}
f(p^j_x) + t_x\\
p^j_y + t_y\\
1\\
\end{bmatrix},
\end{align}
where $\alpha = \rho\cdot\cos(\theta)$ and $\beta = \rho\cdot\sin(\theta)$.

We evaluate the alignment method for all unique image pairs of each plane.
First, we manually annotated the CSP, LV, TCD and HC as shown in the first two rows of \autoref{fig:alignment}.
Each transformation is then evaluated based on the distances of the CSP, LV and TCD centers.
In addition, the alignment of the fetal skull is assessed via the distance of the ellipse centers.
All distances are reported as percent of the respective HC long axis length.
Three baselines are implemented:
First, no alignment (``None'');
second, manually aligning the head orientation via horizontal flipping (``Left-Right'' (LR));
and third, manually aligning the head orientation plus subsequent intensity-based registration (``LR + Intensity'').
For the latter, we compute similarity transformations via the \emph{SimpleElastix} library \cite{Marstal2016}, using the normalized cross-correlation metric with default settings and a maximum of 256 iterations per scale.

\section{Results}
\label{sec:results}
\emph{Salient Landmark Discovery.\quad}
\label{sec:ResLMDisco}
\autoref{fig:examples} shows exemplary results of the salient landmark discovery method.
All shown predicted saliency maps have two peaks: one at the CSP and one at the LV (TV images) or at the cerebellum (TC images).
The cluster labels correctly match the landmarks across images.

\begin{figure}[tb]
{
\tiny
\sffamily
\centering
\def\svgwidth{1.\columnwidth}
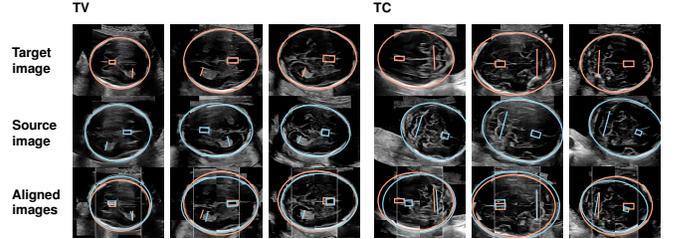
}
\caption{
Exemplary results of the image registration via visually salient landmarks.
The first and second row show target and source images with overlaid annotations of the CSP (box), LV (TV line) and TCD (TC line).
The third row shows the transformed images overlaid with the transformed annotations.
}
\label{fig:alignment}
\end{figure}

\begin{table}[tb]
\caption{
Quantitative results of the image registration with visually salient landmarks and baselines.
The errors for the CSP, LV, cerebellum (``Cereb.'') and HC center are given in percent of the respective HC long axis length.
}
\label{tab:AlignScores}
{
\centering
\small
\aboverulesep=0mm
\belowrulesep=0mm
\begin{tabularx}{\columnwidth}{n t u u u}
\toprule
Plane & Alignment & CSP &LV/Cereb.& HC Center\\
\midrule
\multirow{2}{*}{\STAB{\rotatebox[origin=c]{90}{TV}}}&None&39.3$\;\pm\,$0.3&21.9$\;\pm\,$0.2&15.1$\;\pm\,$0.1\\
&Left-Right (LR)&16.9$\;\pm\,$0.1&\leavevmode\phantom{0}8.9$\;\pm\,$0.1&15.2$\;\pm\,$0.1\\
&LR + Intensity&15.5$\;\pm\,$0.1&\leavevmode\phantom{0}8.2$\;\pm\,$0.1&13.9$\;\pm\,$0.1\\
&Salient LM&\bftab{9.8}$\;\pm\,$0.1&\leavevmode\phantom{0}\bftab{4.1}$\;\pm\,$0.0&\leavevmode\phantom{0}\bftab{7.1}$\;\pm\,$0.0\\
\addlinespace
\multirow{2}{*}{\STAB{\rotatebox[origin=c]{90}{TC}}}&None&58.1$\;\pm\,$0.4&24.8$\;\pm\,$0.2&28.5$\;\pm\,$0.2\\
&Left-Right (LR)&28.4$\;\pm\,$0.2&12.0$\;\pm\,$0.1&24.8$\;\pm\,$0.1\\
&LR + Intensity&27.2$\;\pm\,$0.2&11.6$\;\pm\,$0.1&24.4$\;\pm\,$0.2\\
&Salient LM&\bftab{10.9}$\;\pm\,$0.1&\leavevmode\phantom{0}\bftab{5.7}$\;\pm\,$0.1&\leavevmode\phantom{0}\bftab{6.7}$\;\pm\,$0.0\\
\bottomrule
\end{tabularx}

}
\end{table}

\emph{Application to Image Registration.\quad}
\label{sec:ResLMReg}
After assigning the anatomical structures to the corresponding cluster labels, 88.0\% of the discovered landmarks were near the correct annotated structure (within a radius of 10\% of the HC long axis).
Conversely, 77.1\% of the annotated structures were near a corresponding discovered landmark.
Alignment was performed for 89 (62\%) TV images and 67 (54\%) TC images which had all annotated structures correctly identified.
\autoref{fig:alignment} shows exemplary results and \autoref{tab:AlignScores} shows the corresponding quantitative evaluation.
The alignment errors are consistently lower for salient landmarks compared to the baselines.

\section{Discussion and Conclusion}
The results of \autoref{sec:ResLMDisco} show that the proposed method successfully discovers visually salient landmarks based on predicted human gaze.
While the guidelines define a large set of standard plane criteria via the illustration shown in \autoref{fig:examples}, the landmark discovery method reveals which structures the operators pay attention to in practice.
Specifically, the landmarks correspond to key anatomical structures in the brain, i.e., the LV, cerebellum and CSP.
The CSP itself is not part of any measurement, but it helps the sonographer assess the horizontal orientation of the fetal head and is part of both views \cite{Salomon2011}.
In general, the only prerequisite for applying the landmark discovery method is a set of images from the domain of interest with recorded gaze data in order to train the saliency predictor.

For image registration, the results show that our approach can achieve good alignment without explicit supervision.
The landmarks are successfully matched based on the local features of the saliency prediction CNN.
The intensity-based registration performs significantly worse and only slightly above the trivial ``No align'' and ``Flip'' baselines since intensity-based alignment of ultrasound images is inherently difficult due to noise, shadowing, artifacts and the visibility of maternal anatomies \cite{Che2017}.
The landmark discovery based on visual saliency prediction effectively ignores the irrelevant structures as a human would.
A limitation is that landmark-based alignment is only possible if all necessary landmarks are detected.
Moreover, the quality of alignment may be limited by the affine transform, as visible for the TC plane in \autoref{fig:alignment}, and a non-rigid transformation might yield an improvement.

In conclusion, we have presented a new method to discover visually salient anatomical landmarks
by predicting human gaze.
We have applied the method to fetal neurosonographic images and shown the merit for image alignment compared to intensity-based registration.
Avenues for future work include a comparison of the registration performance to keypoint descriptors (e.g.\ SIFT), and the application of the proposed visually salient landmarks in other areas of radiology, in biological imaging and in cognitive science.

\vspace{4mm}
\begin{small}
\textbf{Acknowledgements.}\quad
We acknowledge the ERC (ERC-ADG-2015 694581, project PULSE), the EPSRC (EP/M013774/1), and the NIHR Oxford Biomedical Research Centre.
\end{small}

\end{document}